\documentclass{article}
\usepackage{spconf,amsmath,epsfig}
\usepackage{amsmath, amsthm, amssymb}
\usepackage{latexsym}
\usepackage{doc}
\usepackage{exscale}
\usepackage{fontenc}
\usepackage{graphicx}

\def \tom {{l,k}}

\title{Audio Classification from Time-Frequency Texture}

\twoauthors {Guoshen Yu}
    {CMAP, Ecole Polytechnique,\\
     91128 Palaiseau Cedex, France}
  {Jean-Jacques Slotine}
   {NSL, Massachusetts Institute of Technology\\
   Cambridge, MA 02139, USA}

\begin{document}
\maketitle

\begin{abstract}
Time-frequency representations of audio signals often resemble texture
images. This paper derives a simple audio classification algorithm
based on treating sound spectrograms as texture images.  The algorithm
is inspired by an earlier visual classification scheme particularly
efficient at classifying textures. While solely based on
time-frequency texture features, the algorithm achieves surprisingly
good performance in musical instrument classification experiments.
\end{abstract}

\section{Introduction}
With the increasing use of multimedia data, the need for automatic
audio signal classification has become an important issue.
Applications such as audio data retrieval and audio file management
have grown in importance~\cite{bregman90, Wold96}.

Finding appropriate features is at the heart of pattern recognition.
For audio classification considerable effort has been dedicated to
investigate relevant features of divers types. Temporal features
such as temporal centroid, auto-correlation~\cite{MPEG7,Brown98},
zero-crossing rate characterize the waveforms in the time domain.
Spectral features such as spectral centroid, width, skewness,
kurtosis, flatness are statistical moments obtained from the
spectrum~\cite{MPEG7,Peeters04}.  MFCCs (mel-frequency cepstral
coefficients) derived from the cepstrum represent the shape of the
spectrum with a few coefficients~\cite{Rabiner93}. Energy
descriptors such as total energy, sub-band energy, harmonic energy
and noise energy~\cite{MPEG7,Peeters04} measure various aspects of
signal power. Harmonic features including fundamental frequency,
noisiness and inharmonicity~\cite{Doval93,MPEG7} reveal the harmonic
properties of the sounds. Perceptual features such as loudness,
shapeness and spread incorporate the human hearing
process~\cite{Zwicker90, Moore1997} to describe the sounds.
Furthermore, feature combination and selection have been shown
useful to improve the classification performance~\cite{Essid06}.

While most features previously studied have an acoustic motivation,
audio signals, in their time-frequency representations, often
present interesting patterns in the visual domain.
Fig.~\ref{fig:spectrograms} shows the spectrograms (short-time
Fourier representations) of solo phrases of eight musical
instruments. Specific patterns can be found repeatedly in the sound
spectrogram of a given instrument, reflecting in part the physics of
sound generation. By contrast, the spectrograms of different
instruments, observed like different textures, can easily be
distinguished from one another. One may thus expect to classify
audio signals in the visual domain by treating their time-frequency
representations as texture images.

In the literature, little attention seems to have been put on audio
classification in the visual domain. To our knowledge, the only work of
this kind is that of Deshpande and his colleges~\cite{Deshpande2001}.
To classify music into three categories (rock, classical, jazz) they
consider the spectrograms and MFCCs of the sounds as visual
patterns. However, the recursive filtering algorithm that they apply
seems not to fully capture the texture-like properties of the audio signal
time-frequency representation, limiting performance.

In this paper, we investigate an audio classification algorithm purely
in the visual domain, with time-frequency representations of audio
signals considered as texture images. Inspired by the recent
biologically-motivated work on object recognition by Poggio, Serre and
their colleagues~\cite{Serre07}, and more specifically on its
variant~\cite{Yu08ICPR} which has been shown to be particularly
efficient for texture classification, we propose a simple feature
extraction scheme based on time-frequency block matching (the
effectiveness of application of time-frequency blocks in audio
processing has been shown in previous
work~\cite{Yu08IEEE,Yu07ICASSP}). Despite its simplicity, the proposed
algorithm relying only on visual texture features achieves
surprisingly good performance in musical instrument classification
experiments.

The idea of treating instrument timbres just as one would treat
visual textures is consistent with basic results in neuroscience,
which emphasize the cortex's anatomical
uniformity~\cite{Mountcastle,Hawkins04} and its functional
plasticity, demonstrated experimentally for the visual and auditory
domains in~\cite{Sur2000}. From that point of view it is not
particularly surprising that some common algorithms may be used in
both vision and audition, particularly as the cochlea generates a
(highly redundant) time-frequency representation of sound.

\section{Algorithm Description}
\begin{figure}[htbp]
\begin{center}
\begin{tabular}{c}
\hspace{-4ex}
\includegraphics[width=9cm]{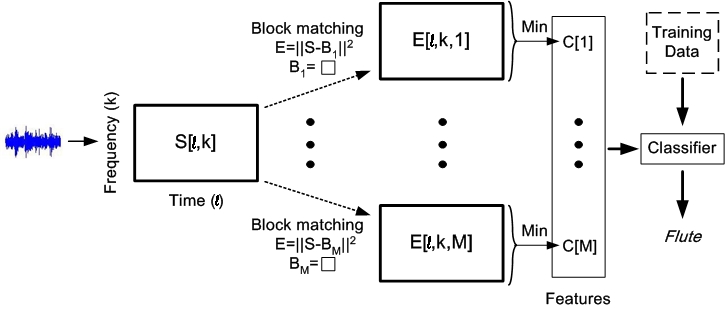}
\end{tabular}
\end{center}
\caption{Algorithm overview.} \label{fig:algo:overview}
\end{figure}

The algorithm consists of three steps, as shown in
Fig.~\ref{fig:algo:overview}. After transforming the signal in
time-frequency representation, feature extraction is performed by
matching the time-frequency plane with a number of time-frequency
blocks previously learned. The minimum matching energy of the blocks
makes a feature vector of the audio signal and is sent to a
classifier.

\subsection{Time-Frequency Representation}

Let us denote an audio signal $f[n]$, $n = 0, 1, \ldots, N-1$. A
time-frequency transform decomposes $f$ over a family of
time-frequency atoms $\{ g_\tom \}_\tom$ where $l$ and $k$ are the
time and frequency (or scale) localization indices. The resulting
coefficients shall be written:
\begin{equation}
F[\tom] = \langle f,g_\tom \rangle = \sum_{n=0}^{N-1} f[n]\,
g_\tom^* [n] \end{equation} where $^*$ denotes the conjugate.
Short-time Fourier transform is most commonly used in audio
processing and recognition~\cite{Yu08IEEE,wavelettour}. Short-time
Fourier atoms can be written: $g_{\tom} [n]= w[n-l u]
\exp\left(\frac{i 2\pi k n}{K}\right) $, where $w[n]$ is a Hanning
window of support size $K$, which is shifted with a step $u \leq K$.
$l$ and $k$ are respectively the integer time and frequency indices
with $0 \leq l < N/u$ and $0 \leq k < K$.

\begin{figure}[htbp]
\begin{center}
\begin{tabular}{cc}
\includegraphics[width=4cm]{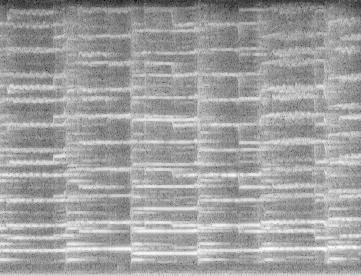} &
\includegraphics[width=4cm]{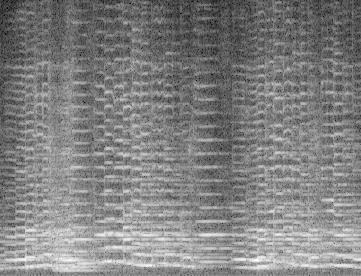} \\
\textbf{violin} & \textbf{cello} \vspace{1ex}\\
\includegraphics[width=4cm]{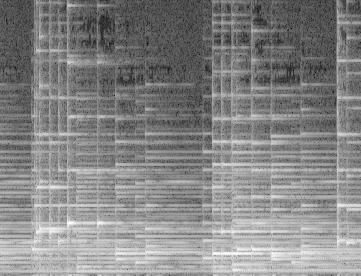} &
\includegraphics[width=4cm]{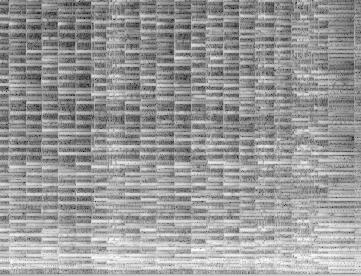} \\
\textbf{Piano} & \textbf{Harpsichord} \vspace{1ex}\\
\includegraphics[width=4cm]{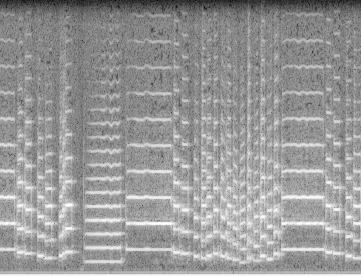} &
\includegraphics[width=4cm]{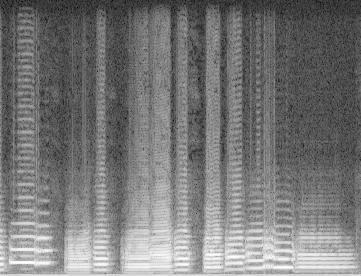} \\
\textbf{Trumpet} & \textbf{Tuba} \vspace{1ex}\\
\includegraphics[width=4cm]{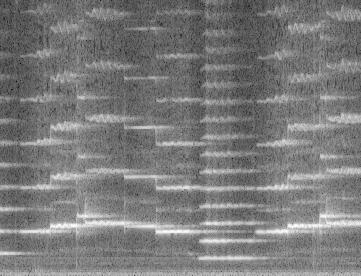} &
\includegraphics[width=4cm]{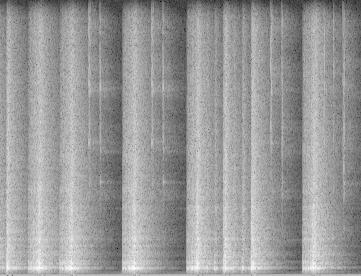} \\
\textbf{Flute} & \textbf{Drum} \\
\end{tabular}
\end{center}
\vspace{-2ex} \caption{Log-spectrograms of solo phrases of different
musical instruments.} \label{fig:spectrograms} \vspace{-2ex}
\end{figure}

The time-frequency representation provides a good domain for audio
\textit{classification} for several reasons. First, of course, as the
time-frequency transform is invertible, the time-frequency
representation contains complete information of the audio signal. More
importantly, the texture-like time-frequency representations usually
contain distinctive patterns that capture different characteristics of
the audio signals. Let us take the spectrograms of sounds of musical
instruments as illustrated in Fig.~\ref{fig:spectrograms} for
example. Trumpet sounds often contain clear onsets and stable
harmonics, resulting in clean vertical and horizontal structures in
the time-frequency plane. Piano recordings are also rich in clear
onsets and stable harmonics, but they contain more chords and the
tones tend to transit fluidly, making the vertical and horizontal
time-frequency structures denser. Flute pieces are usually soft and
smooth. Their time-frequency representations contain hardly any
vertical structures, and the horizontal structures include rapid
vibrations. Such textural properties can be easily learned without
explicit detailed analysis of the corresponding patterns.

As human perception of sound intensity is
logarithmic~\cite{Zwicker90}, the classification is based on
log-spectrogram
\begin{equation}
S[\tom]=\log|F[\tom]|.
\end{equation}

\subsection{Feature Extraction}
Assume that one has learned $M$ time-frequency blocks $B_m$ of size
$W_m \times L_m$, each block containing some time-frequency
structures of audio signals of various types. To characterize an
audio signal, the algorithm first matches its log-spectrogram $S$
with the sliding blocks $B_m$, $\forall m=1,\ldots,M$,
\begin{equation}
\label{eqn:block:matching} E[l,k,m] = \frac{\sum_{i=1}^{W_m}
\sum_{j=1}^{L_m} |S[l+i-1,k+j-1]-B_m[i,j]|^2}{W_m L_m}.
\end{equation}
$E[l,k,m]$ measures the degree of resemblance between the patch
$B_m$ and log-spectrogram $S$ at position $[l,k]$. A minimum
operation is then performed on the map $E[l,k,m]$ to extract the
highest degree of resemblance locally between $S$ and $B_m$:
\begin{equation}
C[m] = \min_{l,k} E[l,k,m].
\end{equation}
The coefficients $C[m]$, $m=1,\ldots,M$, are time-frequency
translation invariant. They constitute a feature vector $\{C[m]\}$
of size $M$ of the audio signal. Let us note that the block matching
operation (\ref{eqn:block:matching}) can be implemented fast by
convolution.

The feature coefficient $C[m]$ is expected to be discriminative if
the time-frequency block $B_m$ contains some salient time-frequency
structures. In this paper, we apply a simple random sampling
strategy to learn the blocks as in~\cite{Serre07,Yu08ICPR}: each
block is extracted at a random position from the log-spectrogram $S$
of a randomly selected training audio sample. Blocks of various
sizes are applied to capture time-frequency structures at different
orientations and scales~\cite{Yu08IEEE}. Since audio time-frequency
representations are rather stationary images and often contain
repetitive patterns, the random sampling learning is particularly
efficient. Patterns that appear with high probability are likely to
be learned.

\subsection{Classification}
The classification uses the minimum block matching energy $C$
coefficients as features. While various classifiers such as SVMs can
be used, a simple and robust nearest neighbor classifier will be
applied in the experiments.

\section{Experiments and Results}
The audio classification scheme is evaluated through musical
instrument recognition. Solo phrases of eight instruments from
different families, namely flute, trumpet, tuba, violin, cello,
harpsichord, piano and drum, were considered. Multiple instruments
from the same family, violin and cello for example, were used to avoid
over-simplification of the problem.

To prepare the experiments, great effort has been dedicated to
collect data from divers sources with enough variation, as few
databases are publicly available.  Sound samples were mainly
excerpted from classical music CD recordings of personal
collections. A few were collected from internet. For each instrument
at least 822-second sounds were assembled from more than 11
recordings, as summarized in Table~\ref{tab:data}. All recordings
were segmented into non-overlapping excerpts of 5 seconds. 50
excerpts (250 seconds) per instrument are randomly selected to
construct respectively the training and test data sets. The training
and test data did not contain certainly the same excerpts. In order
to avoid bias, \textit{excerpts} from the \textit{same recording}
were never included in both the training set and the test set.

\begin{table}[t]
\begin{center}
\begin{tabular}{|c|c|c|c|c|c|c|c|c|}
\hline & {\small Vio.} & {\small Cel.} & {\small Pia.} & {\small
Hps.} & {\small Tru.} & {\small Tuba} & {\small Flu.} &
{\small Drum} \\
\hline {\small Rec.} & {\small27} & {\small35} & {\small31} & {\small68} & {\small11} & {\small15} & {\small12} & {\small22} \\
\hline {\small Time} & {\small7505} & {\small7317} & {\small6565} & {\small 11036} & {\small822} & {\small1504} & {\small2896} & {\small2024} \\
\hline
\end{tabular}
\end{center}
\vspace{-2ex} \caption{Sound database. \textit{Rec} and
\textit{Time} are the number of recordings and the total time
(second). Musical instruments from left to right: violin, cello,
piano, harpsichord, trumpet, tuba, flute and drum.} \label{tab:data}
\vspace{-3ex}
\end{table}

Human sound recognition performance seems not degrade if the signals
are sampled at 11000 Hz. Therefore signals were down-sampled to
11025 Hz to limit the computational load. Half overlapping Hanning
windows of length 50 ms were applied in the short-time Fourier
transform. Time-frequency blocks of seven sizes $16 \times 16$, $16
\times 8$ and $8 \times 16$, $8 \times 8$, $8 \times 4$ and $4
\times 8$ and $4 \times 4$ that cover time-frequency areas of size
from $640 \textrm{Hz} \times 800 \textrm{ms}$ to $160 \textrm{Hz}
\times 200 \textrm{ms}$ were simultaneously used, same number for
each, to capture time-frequency structures at different orientations
and scales. The classifier was a simple nearest neighbor
classification algorithm.

\begin{figure}[htbp]
\begin{center}
\begin{tabular}{c}
\includegraphics[width=7cm]{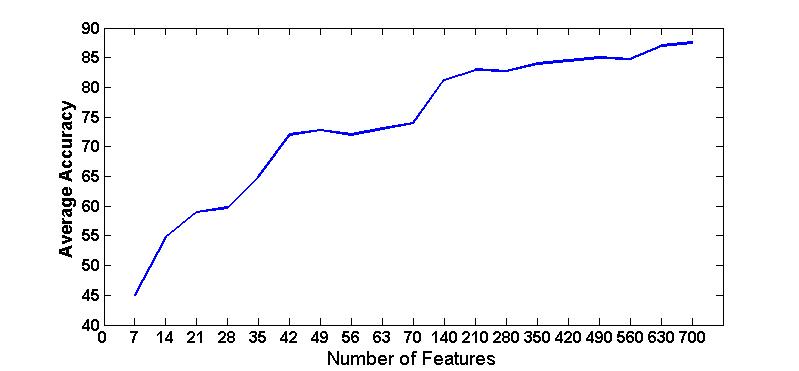}
\end{tabular}
\end{center}
\caption{Average accuracy versus number of features.}
\label{fig:accuracy:numfeatures}
\end{figure}

Fig.~\ref{fig:accuracy:numfeatures} plots the average accuracy
achieved by the algorithm in function of the number of features
(which is seven times the number of blocks per block size). The
performance rises rapidly to a reasonably good accuracy of 80\% when
the number of features increases to about 140. The accuracy
continues to improve slowly thereafter and becomes stable at about
85\%, very satisfactory, after the number of features goes over 350.
Although this number of visual features looks much bigger than the
number of carefully designed classical acoustic features (about 20)
commonly used in literature~\cite{Guo03, Essid06}, their computation
is uniform and very fast.

The confusion matrix in Table~\ref{tab:confusion} reveals the
classification details (with 420 features) of each instrument. The
highest confusion arrived between the harpsichord and the piano
which can produce very similar sounds. Other pairs of instruments
that produce potentially sounds of similar nature such as flute and
violin were occasionally confused. Some trumpet excerpts were
confused with violin and flute --- these excerpts were found rather
soft and contained mostly harmonics. The drum that is most distinct
from the others had the lowest confusion rate. The average accuracy
is 85.5\%.

\begin{table}[htbp]
\begin{center}
\begin{tabular}{|c|c|c|c|c|c|c|c|c|}
\hline & {\small Vio.} & {\small Cel.} & {\small Pia.} & {\small
Hps.} & {\small Tru.} & {\small Tuba} & {\small Flu.} &
{\small Drum} \\
\hline{\small Vio.} & 94&  0&  0&  0&  2&   0&   4& 0\\
\hline{\small Cel.} &  0& 84&  6&  10&  0&   0&   0& 0\\
\hline{\small Pia.} &  0&  0& 86& 8&  6&   0&   0& 0\\
\hline{\small Hps.} &  0&  0&  26& 74&  0&   0&   0& 0\\
\hline{\small Tru.} &  8&  2&  2&  0& 80&   0&   8& 0\\
\hline{\small Tuba} &  2&  4&  2&  0&  0& 90&   0& 2\\
\hline{\small Flu.} &  6&  0&  0&  0&  0&   0&  94& 0\\
\hline{\small Drum} &  0&  0&  0&  0&  0&   2&  0& 98\\
\hline
\end{tabular}
\end{center}
\vspace{-2ex} \caption{Confusion matrix. Each entry is the
 rate that the row instrument is classified as the column
instrument. Musical instruments from top to bottom, left to right:
violin, cello, piano, harpsichord, trumpet, tuba, flute and drum.}
\label{tab:confusion}
\end{table}

\section{Conclusion and Future Work}

An audio classification algorithm is proposed, with spectrograms of
sounds treated as texture images.  The algorithm is inspired by an
earlier biologically-motivated visual classification scheme,
particularly efficient at classifying textures. In experiments, this
simple algorithm relying purely on time-frequency texture features
achieves surprisingly good performance at musical instrument
classification.

In future work, such image features could be combined with more
classical acoustic features. In particular, the still largely unsolved
problem of instrument separation in polyphonic music may be simplified
using this new tool. \\

\vspace{0ex} {\small {\noindent \textbf{Acknowledgements:} We are
grateful to Emmanuel Bacry, Jean-Baptiste Bellet, Laurent Duvernet,
St\'ephane Mallat, Sonia Rubinsky and Mingyu Xu for their
contribution to the audio data collection.}}

\bibliographystyle{plain}

\end{document}